\def\year{2016}
\documentclass[letterpaper]{article}
\usepackage{aaai16}
\usepackage{times}
\usepackage{helvet}
\usepackage{courier}
\usepackage{slashbox}
\usepackage{graphicx}
\usepackage{csquotes}
\usepackage{amsmath}
\usepackage{amssymb}
\usepackage{algorithm}
\usepackage{algorithmic}
\usepackage{hyperref}

\def\W{{\bf{W}}}

\def\etal{{ et al.}}

\begin{document}
%
\title{Look, Listen and Learn - A Multimodal LSTM for Speaker Identification}
\author{
Jimmy Ren$^{1}$ \hspace{0.03in} Yongtao Hu$^{2}$ \hspace{0.03in} Yu-Wing Tai$^{1}$ \hspace{0.03in} Chuan Wang$^{2}$ \hspace{0.03in}
Li Xu$^{1}$ \hspace{0.03in} Wenxiu Sun$^{1}$ \hspace{0.03in} Qiong Yan$^{1}$\\\\
SenseTime Group Limited$^{1}$\\
\{rensijie, yuwing, xuli, sunwenxiu, yanqiong\}@sensetime.com\\
The University of Hong Kong$^{2}$\\
\{herohuyongtao, wangchuan2400\}@gmail.com\\
Project page: http://www.deeplearning.cc/mmlstm
}
\maketitle
\begin{abstract}
\begin{quote}
Speaker identification refers to the task of localizing the face of a person who has the same identity as the ongoing voice in a video. This task not only requires collective perception over both visual and auditory signals, the robustness to handle severe quality degradations and unconstrained content variations are also indispensable. In this paper, we describe a novel multimodal Long Short-Term Memory (LSTM) architecture which seamlessly unifies both visual and auditory modalities from the beginning of each sequence input. The key idea is to extend the conventional LSTM by not only sharing weights across time steps, but also sharing weights across modalities. We show that modeling the temporal dependency across face and voice can significantly improve the robustness to content quality degradations and variations. We also found that our multimodal LSTM is robustness to distractors, namely the non-speaking identities. We applied our multimodal LSTM to The Big Bang Theory dataset and showed that our system outperforms the state-of-the-art systems in speaker identification with lower false alarm rate and higher recognition accuracy.
\end{quote}
\end{abstract}

\noindent Speaker identification is one of the most important building blocks in many intelligent video processing systems such as video conferencing, video summarization and video surveillance, etc. It aims to localize the face of the speaker associated with the ongoing voices. To achieve this task, collective perception over both visual and auditory signals is indispensable.

\begin{figure}[t]
  \centering
  \includegraphics[width=1.1\linewidth]{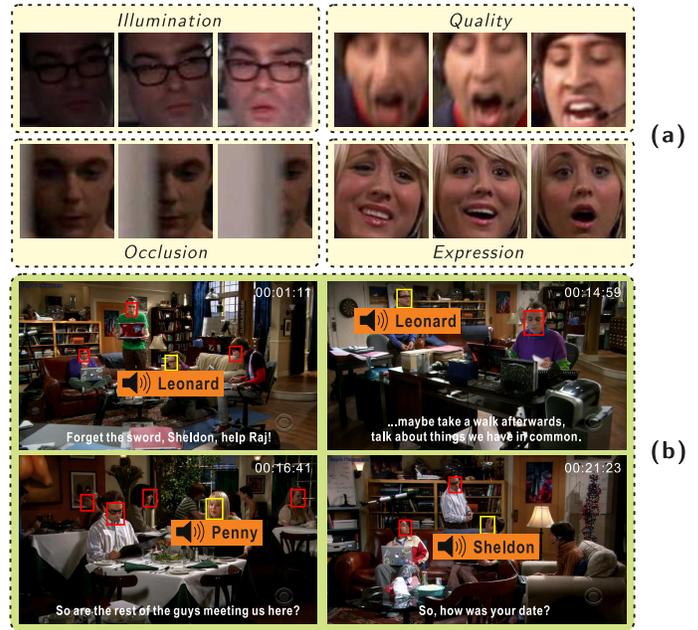}\vspace{-0.1in}
  \caption{(a) Face sequence with different kinds of degradations and variations. Using the previous CNN methods cannot recognize the speakers correctly. In contrast, the speakers can be successfully recognized by our LSTM in both single-modal and multimodal settings. (b) Our multimodal LSTM is robust to both image degradation and distractors. Yellow bounding boxes are the speakers. Red bounding boxes are the non-speakers, the distractors.} \vspace{-0.1in}
  \label{fig:fig1} 
\end{figure}

In the past few years, we observe the rapid advances in face recognition and speech recognition respectively by using Convolutional Neural Networks (CNN) \cite{Schroff15,Sun14} and Recurrent Neural Networks (RNN) \cite{Graves13,Hannum14}. Notwithstanding the recent ground breaking results in processing facial and auditory data, speaker identification (figure \ref{fig:fig1}b) remains challenging for the following reasons. First, severe quality degradations (e.g. blur and occlusion) and unconstrained content variations (e.g. illumination and expression) in real-life videos are not uncommon. These effects significantly degrade the performance of many existing CNN based methods. Second, the state-of-the-art convolutional network based face model is trained with still images. Its application in sequential data as well as its connection to recurrent networks is less explored and it is not straightforward to extend CNN based methods to the multimodal learning setting. Third, a practical system should be robust enough to reject distractors, the faces of non-speaking persons indicated by the red bounding boxes in figure \ref{fig:fig1}b, which adds additional challenges to the task.

Despite the rich potential of both CNN and RNN for facial and auditory data, it is still unclear if these two networks can be simultaneously and effectively adopted in the context of multimodal learning. In this paper, we proposed a novel multimodal Long Short-Term Memory (LSTM) \cite{Hochreiter97}, a specialized type of Recurrent Neural Network, to address this problem. We show that modeling temporal dependency for facial sequences by LSTM performs much better than CNN based methods in terms of robustness to image degradations. More importantly, by sharing weights not only across time steps but also across different modalities, we can seamlessly unify both visual and auditory modalities from the beginning of each sequence input, which significantly improves the robustness to distractors. This is because the cross-modal shared weights can learn to capture temporal correlation between face and voice sequences. Note that our multimodal LSTM did not assume the the random variables from different modal are correlated. Instead, our multimodal LSTM is capable to learn such correlation if there is any temporal correlation across different modal in their sequences. In the speaker identification task, we believe such temporal correlation exists in the respective face and voice sequences. In this work, we assume the face of the speaker appears in a video when they speak, and there is only one speaker at the same time during the voice over. Multiple speakers in the same video clip can be identified as far as their speaks do not overlap.

To our knowledge, our paper is the first attempt in modeling long-term dependencies over multimodal high-level features which demonstrates robustness to both distractors and image degradation. We applied our model to The Big Bang Theory dataset and showed that our system outperformed the state-of-the-art systems in recognition accuracy and with lower false alarm rate. 

The contributions of this paper are as follows. 
\begin{itemize}
\item We proposed a novel LSTM architecture which enables multimodal learning of sequence data in a unified model. Both temporal dependency within each modality and temporal correlation across modalities can be automatically learned from data.
\item We empirically showed that cross-modality weight sharing in LSTM simultaneously improves the precision of classification and the robustness to distractors.
\item We successfully applied our method in a real-world multimodal classification task and the resulting system outperformed the state-of-the-art. The dataset and our implementations are both publicly available. 
\end{itemize}

\section{Related Work}
\noindent Many recent studies have reported the success of using deep CNN in face related tasks. The pioneering work by \cite{Taigman14} proposed a very deep CNN architecture together with an alignment technique to perform face verification which achieved near human-level performance. Inspired by GoogLeNet \cite{Szegedy15}, Sun et al. \citeyear{Sun14b} used a very deep CNN network with multiple levels of supervision, which surpassed human-level face verification performance in the LFW dataset \cite{Huang13}. The recent advance in this field \cite{Schroff15} pushed the performance even further. In face detection, the state-of-the-art results were also achieved by CNN based models \cite{Yang15,Li15}. For other face related tasks such as face landmark detection and face attribute recognition \cite{Zhang15,Zhang15b}, CNN based models were also widely adopted. 

The revived interest on RNN is mainly attributed to its recent success in many practical applications such as language modeling \cite{Kiros15}, speech recognition \cite{Chorowski15,Graves13}, machine translation \cite{Sutskever14,Jean15}, conversation modeling \cite{Shang15} to name a few. Among many variants of RNNs, LSTM is arguably one of the most widely used model. LSTM is a type of RNN in which the memory cells are carefully designed to store useful information to model long term dependency in sequential data \cite{Hochreiter97}. Other than supervised learning, LSTM is also used in recent work in image generation \cite{Theis15,Gregor15}, demonstrating its capability of modeling  statistical dependencies of imagery data.

In terms of the sequence learning problem across multiple modalities, LSTM based models were actively used in recent image caption generation studies \cite{Donahue15,Karpathy15,Xu15}. One common characteristic of these techniques is that CNN was used to extract the feature sequences in an image and LSTM was used to generate the corresponding text sequences. Our paper is related to this group of studies in a way that more than one modalities are involved in the learning process. However, our goal is not to generate sequences in an alternative domain but to collectively learn useful knowledge from sequence data of multiple domains. Perhaps the most closely related previous studies to our work are from \cite{Srivastava12} and \cite{Ngiam11}. Unlike these papers, we focused on high-level multimodal learning which explicitly models the temporal correlation of high-level features rather than raw inputs between different modalities. This not only provided a channel to effectively transfer the recent success of deep CNN to the multimodal learning context, the resulting efficient implementation can be directly adopted in video processing as well. We also investigated the robustness to distractors and input quality which is not considered in the previous studies. The closely related papers in multimedia speaker identification are \cite{Bauml2013,hu2015deep}, and \cite{Tapaswi2012}. However, they did not explicitly model face sequences and the interplay between face and voice sequences.

\section{LSTM - Single VS. Multi-Modal}
\begin{figure}[t]
  \centering
  \includegraphics[width=1.0\linewidth]{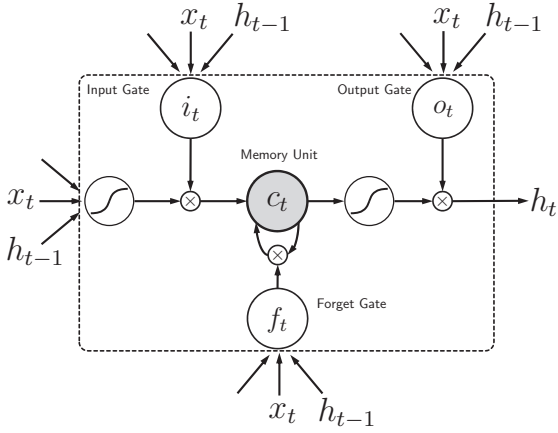}\vspace{-0.1in}
  \caption{Memory Cell of Single-modal LSTM.}\vspace{-0.15in}
  \label{fig:old_lstm} 
\end{figure}

\paragraph{Single Modal LSTM} A regular LSTM network contains a number of memory cells within which the multiplicative gate units and the self-recurrent units are the two fundamental building blocks \cite{Hochreiter97}. For a brief revision, equations (\ref{eq:old_input}), (\ref{eq:old_forget_gate}) and (\ref{eq:old_recurrent_unit}) formally describe the memory input, the forget gate and the recurrent units of a regular LSTM in the forward pass. The input gate $i_t$ and the output gate $o_t$ in a regular LSTM resemble the forget gate in the forward pass. Figure \ref{fig:old_lstm} shows a pictorial illustration of a regular LSTM model.
\begin{eqnarray}
\hspace{-0.1in}g_t\hspace{-0.1in} &=& \hspace{-0.1in} \varphi(\W_{xg}*X_t+\W_{hg}*h_{t-1}+b_g), \label{eq:old_input} \\
\hspace{-0.1in}i_t\hspace{-0.1in} &=& \hspace{-0.1in} \sigma(\W_{xi}*X_t+\W_{hi}*h_{t-1}+b_i), \label{eq:old_input_gate} \\
\hspace{-0.1in}f_t\hspace{-0.1in} &=& \hspace{-0.1in} \sigma(\W_{xf}*X_t+\W_{hf}*h_{t-1}+b_f), \label{eq:old_forget_gate} \\
\hspace{-0.1in}o_t\hspace{-0.1in} &=& \hspace{-0.1in} \sigma(\W_{xo}*X_t+\W_{ho}*h_{t-1}+b_o), \label{eq:old_output_gate} \\
\hspace{-0.1in}C_t\hspace{-0.1in} &=& \hspace{-0.1in} f_t \odot C_{t-1}+ i_t \odot g_t, \label{eq:old_recurrent_unit} \\
\hspace{-0.1in}y_t\hspace{-0.1in} &=& \hspace{-0.1in} softmax(\W_y*h_t). \label{eq:old_output}
\end{eqnarray}

In (\ref{eq:old_input}) and (\ref{eq:old_forget_gate}), $X$ is an input sequence where $X_t$ is an element of the sequence at time $t$, $h_{t-1}$ is the output of the memory cell at time $t-1$. $\W_{xg}, \W_{xf}, \W_{hg}, \W_{hf}$ are distinct weight matrices, and $b_g$ and $b_f$ are bias terms respectively. $\varphi$ and $\sigma$ are nonlinear functions where $\varphi$ denotes a $tanh$ function and $\sigma$ denotes a $sigmoid$ function. In (\ref{eq:old_recurrent_unit}), $\odot$ denotes an element-wise multiplication, $i_t$ is the input gate at the time step $t$, and $C_{t-1}$ is the memory unit at the time step $t-1$. The memory unit at time step $t$ is therefore generated by the collective gating of the input gate and the forget gate. In equation (\ref{eq:old_output}), the memory cell output of the current time step is multiplied by $\W_y$ and then transformed by the $softmax$ function to compute the model output $y_t$ at time $t$.

Generally speaking, the reason that LSTM is able to model long-term dependencies in sequential data is because $C_t$ at each time step can selectively ``remember'' (store) or ``forget'' (erase) past information which is modelled by the multiplicative gating operation. More importantly, the strategy to open or to close the gates is data driven which is automatically learned from training data. This information is captured by the trainable weights $\W$, including $\W_{hf}, \W_{xf}$ and so on, rather than hand-crafted. Because $\W$ are shared across time steps, this endows LSTM the power to explicitly model temporal relationships over the entire sequence.

\vspace{-0.15in}
\paragraph{Simple extensions of Single Modal LSTM}
In order to deal with data from different domains, perhaps the most straightforward method is to incorporate them into a single network by concatenating the data directly to produce a bigger $X$. However, this approach is problematic because the multimodal property of the inputs are completely ignored and the model does not have any explicit mechanism to model the correlation across modalites. Though some correlations may be weakly captured by the trained weights, a critical weakness is that it is incapable to handle distractors. In particular, when the face of a person A is combined with the voice of a person B, the model is confused and would fail to generate a meaningful label. Although it may be possible to put all the distractors to a single class and let the model to distinguish the speaker and the distractors automatically, this method performs much worse than our solution in practice. The major difficulty is that distractors share too many features with regular examples when organize the inputs in this way.

Another solution is to treat data from different domains completely independent. Namely, we can use multiple LSTMs in parallel and then merge the output labels at the highest layer using a voting mechanism. The advantage of this approach is that the two separate memory units can be trained to store useful information explicitly for each domain. But the weakness is that the interaction across modalities only happens at the highest level during the labelling process. The cross-model correlation is therefore very difficult, if not entirely impossible, to be encoded into the weights through the learning process. Thus, the robustness to distractors relies heavily on the voting stage where some of the temporal correlations may have already been washed out in the independent forward pass.

\begin{figure}[t]
  \centering
  \includegraphics[width=1.0\linewidth]{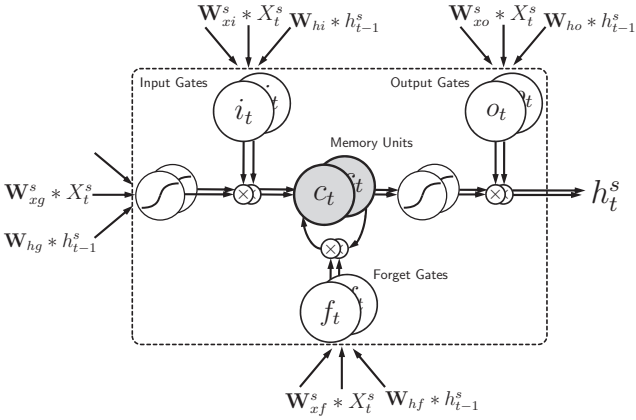}\vspace{-0.1in}
  \caption{Memory Cell of Multimodal LSTM.}\vspace{-0.15in}
  \label{fig:fig2} 
\end{figure}

\vspace{-0.15in}
\paragraph{Multimodal LSTM}
Compared with the straightforward solutions, we want to develop a new multimodal LSTM which can explicitly model the long-term dependencies both within the same modality and across modalities in a single multimodal LSTM.
Instead of merging input data at pre-processing stage, or merging labels at post-processing stage, our key idea is to selectively share weights across different modalities during the forward pass. This is similar to the weight sharing in time domain in regular LSTM, but we do not share memory units for each modality within the memory cell. The modifications are illustrated in figure \ref{fig:fig2} and formally expressed in the following equations.
\begin{eqnarray}
\hspace{-0.18in}g_t^s\hspace{-0.1in} &=& \hspace{-0.1in} \varphi(\W_{xg}^s*X_t^s+\W_{hg}*h_{t-1}^s+b_g^s), ~~ s = 1~to~n, \label{eq:new_input} \\
\hspace{-0.18in}i_t^s\hspace{-0.1in} &=& \hspace{-0.1in} \sigma(\W_{xi}^s*X_t^s+\W_{hi}*h_{t-1}^s+b_i^s), ~~ s = 1~to~n, \label{eq:new_input_gate} \\
\hspace{-0.18in}f_t^s\hspace{-0.1in} &=& \hspace{-0.1in} \sigma(\W_{xf}^s*X_t^s+\W_{hf}*h_{t-1}^s+b_f^s), ~~ s = 1~to~n,\label{eq:new_forget_gate} \\
\hspace{-0.18in}o_t^s\hspace{-0.1in} &=& \hspace{-0.1in} \sigma(\W_{xo}^s*X_t^s+\W_{ho}*h_{t-1}^s+b_o^s), ~~ s = 1~to~n, \label{eq:new_output_gate} \\
\hspace{-0.18in}C_t^s\hspace{-0.1in} &=& \hspace{-0.1in} f_t^s \odot C_{t-1}^s+i_t^s \odot g_t^s, ~~ s = 1~to~n, \label{eq:new_recurrent_unit} \\
\hspace{-0.18in}h_t^s\hspace{-0.1in} &=& \hspace{-0.1in} o_t^s \odot \varphi(C_t^s), ~~ s = 1~to~n, \label{eq:new_cell_output} \\
\hspace{-0.18in}y_t^s\hspace{-0.1in} &=& \hspace{-0.1in} softmax(\W_y*h_t^s), ~~ s = 1~to~n. \label{eq:new_output}
\end{eqnarray}

Keeping the gating mechanism the same as the regular LSTM, the equations from (\ref{eq:new_input}) to (\ref{eq:new_output}) describe a new cross-modal weight sharing scheme in the memory cell. The superscript $^s$ indexes each modality in the input sequences. $n$ is the total number of modalities in input data, where $n=2$ in the speaker identification task. The model is general enough to deal with the tasks with $n>2$. $X_t^s$ is the input sequence at time $t$ for modality $s$. Therefore, the weights with superscript $^s$ (e.g. $\W_{xg}^s$) are NOT shared across modalities but only across time steps, the other weights without the superscript (e.g. $\W_{hg}$) are shared across both modalities and time steps. Specifically, the weights associated with the inputs $X_t^s$ are not shared across modalities. The reasons are two-fold. First, we would like to learn a specialized mapping, separately for each modality, from its input space to a new space, where multimodal learning is ensured by the shared weights. Second, specialized weights are preferred to reconcile the dimension difference between different modalities, which avoids a complex implementation.

Along with the transform associated with $X_t^s$, the output of the memory cell from the previous time step $h_{t-1}^s$ also need to go through a transform in producing $g_t^s$ as well as all other gates. The weights to perform this transform, $\W_{hg}, \W_{hi}, \W_{hf}$ and $\W_{ho}$ are shared across the modalities. With these new weight definitions, the separately transformed data by the four $\W^s$ is essentially interconnected from $g_t^s$ all the way to the memory cell output $h_t^s$.

The key insight is that while it is preferable for each modality to have its own information flow because it enables a more focused learning objective with which it is easier to learn the long-term temporal dependency within the modality, we also make such objective correlated and essentially constrained by what happens in the rest of the modalities. More specifically, in forming the forget gate $f_t^s$ for $s=1$, it not only relates to $h_{t-1}^{s=1}$ but also constrained by $h_{t-1}^{s=2}$, ... , $h_{t-1}^{s=n}$ because $\W_{hf}$ is shared among them. Provided that the weights $\W_{hg}, \W_{hi}, \W_{hf}$ and $\W_{ho}$ are also shared across time steps, they play the vital role of capturing the temporal correlation across different modalities.

Another important property of the proposed model is that the memory unit $C$ is NOT shared among modalities. The rationale is that the weights have the job to capture intramodal as well as intermodal relationships, therefore placing them in a single memory unit provides much less flexibility on what can be stored or forgotten. Given all the gates are formed in a multimodal fashion, the insight of such design is that we should not hand-craft the decision on what intramodal/intermodal relationships should be stored or forgotten but to give the model enough flexibility to learn it from data. The bias terms are not shared across modalities neither to increase this flexibility.

Likewise, the network output at each time step $y_t^s$ does not relate to its own modality. Whether we should use $\W_y$ or $\W_y^s$ to transform $h_t^s$ before sending the outputs to the softmax function is not a straightforward decision. We resort to our experiment to address this issue.

\vspace{-0.15in}
\paragraph{High-Level Feature VS. Raw Inputs}
One of the most important reasons why CNN is attractive is because it is an end to end feature learning process. Previous studies \cite{Razavian14,Oquab14} have discovered that a successful CNN for a classification task also produces high-level features which are general enough to be used in a variety of tasks. This finding inspired a few recent work on image captioning \cite{Xu15,Karpathy15} where the high-level CNN features over an image sequence were extracted, and a RNN is learned on top of the extracted CNN features to perform more sophisticated tasks. Such approach is very effective to bridge the effort and success in CNN to the field of sequence modeling. We would like to extend this type of attempt to multimodal sequence learning.

\vspace{-0.15in}
\paragraph{Implementation}
In order to maximize the flexibility of our investigation and efficiently work with different variants of network architecture and working environments (e.g. Linux and Windows), we did not implement the multimodal LSTM using any third-party deep learning packages. Instead, we used MATLAB and its GPU functions in the parallel computing toolbox to build our own LSTM from scratch. Our implementation is vectorized \cite{Ren14_aaai} and very efficient in training both single-modal and multimodal LSTM described in this paper.

\section{Experiments}
Three experiments were carefully designed to test the robustness and the applicability of our proposed model.

\vspace{-0.15in}
\paragraph{Dataset overview}
We chose the TV-series The Big Bang Theory (BBT) as our data source. It has been shown that the speaker identification task over BBT is a very challenging multimodal learning problem due to various kinds of image degradation and the high variations on faces in the videos \cite{Bauml2013,hu2015deep,Tapaswi2012}. During data collection, we ran face detection and extracted all the faces in six episodes in the first season and another six episodes in the second season of BBT. We manually annotated the faces for the five leading characters, i.e. \textit{Sheldon}, \textit{Leonard}, \textit{Howard}, \textit{Raj} and \textit{Penny}. In total, we have more than 310,000 consecutively annotated face images for the five characters. For audio data, we utilized the pre-annotated subtitles and only extracted the audio segments corresponding to speeches. Data from the second season was used in training and data from the first season was used in testing for all the experiments reported below.

\vspace{-0.15in}
\paragraph{Feature extraction}
To ensure the usability of the resulting system, we adopted 0.5 second as the time window of all the sequence data including both face and audio. For feature extraction for faces, we adopted a CNN architecture resembles the one in \cite{KrizhevskySH12} and trained a classifier using the data reported in the next section. The activations of the last fully connected layer was used as the high-level feature for face. We also run principle component analysis (PCA) on all the extracted face features to reduce the dimensionality to the level comparable to audio features. By keeping 85\% of the principle components, we obtained a 53-dimension feature vector for each face in the video. The video is 24 frames per second, therefore there are 12 consecutive faces within each face sequence. For audio, we used the mel-frequency cepstral coefficients (MFCC) features \cite{Sahidullah12}. Following \cite{hu2015deep} we extracted the 25d MFCC features in a sliding window of 20 milliseconds with stride of 10 milliseconds, thus gives us 25$\times$49 MFCC features.

\subsection{LSTM for Face Sequences}
Our first task is to investigate the extend to which modeling temporal dependency of high-level CNN features improves the robustness to quality degradation and face variation. The reasons that we would like to investigate this manner is two-fold. First, though it was showed that RNN can be successfully used in speech recognition to improve the robustness to noise \cite{Graves13}, it was not clear from the literature whether similar principle applies for high-level image features. On the other hand, we would like to clearly measure the extend to which this approach works for faces because this is an important cornerstone for the rest of the experiments.

\vspace{-0.15in}
\paragraph{Data}
Only face data in the aforementioned data set was used in the experiment. We randomly sampled 40,000 face sequences from the training face images and another 40,000 face sequences in the test face images. Note that each sequence was extracted according to the temporal order in the data, however, we did not guarantee the sequence are strictly from one subtitle segment. This injected more variations in the sequence.

\vspace{-0.15in}
\paragraph{Procedure and Results}
Three methods were compared in this experiment. The first method was to use a CNN to directly classify each frame in the sequence. This CNN is the same one as used in the feature extraction for our LSTM. In our setting, the CNN will output 12 labels (12 probability distributions) for each face sequence. Then the output probabilities were averaged to compute the final label for this sequence. The second method used the same CNN to extract features for each frame and reduced the dimensionality using PCA as described in the last section. Then we used a SVM with RBF kernel to classify each feature followed by the same averaging processing before outputting the label. We used the single-modal LSTM (see figure \ref{fig:old_lstm}) with one hidden memory cell layer to train a sequential classifier. The dimensionality of the hidden layer activation is 512. In our setting, this LSTM contains 12 time steps with 12 identical supervision signals (labels). During the testing, we only look at the last output in the whole output sequence.

The results were reported in table \ref{table:svm-compare}. We can see that the two CNN alone approaches delivered very similar results, acknowledging the high representative powerful of the CNN features reported in the previous studies. The accuracy of the CNN+SVM approach slightly outperformed the CNN alone approach. This is reasonable because SVM with RBF kernel may classify the data better than the last layer of CNN. The performance of LSTM is significantly higher than the other two. By looking at the correctly classified face sequences which were failed in the other two methods, we can see that the LSTM is more robust to a number of image degradations and variations. This is illustrated in figure \ref{fig:fig1}a.

\begin{table}
\begin{center}
\caption{Face sequence classification accuracy of different algorithms.}\vspace{-0.05in}
\label{table:svm-compare}
\scalebox{0.82}{
\begin{tabular}{|l|c|}
\hline
\textsc{Algorithms} & \textsc{Accuracy ($\%$)} \\
\hline\hline
CNN & 92.33 \\
CNN+SVM & 92.42 \\
LSTM & 95.61 \\
\hline
\end{tabular}
}
\end{center}\vspace{-0.15in}
\end{table}

\subsection{Comparison among Multimodal LSTMs}
By the results from the first experiment, there is a reason to believe that performing multimodal learning of face and audio in temporal domain, if do it correctly, has the potential to perform better in speaker identification task. Therefore, the aim of the second experiment was to examine the extend to which the multimodal LSTM benefits the speaker identification performance. Multiple aforementioned multimodal LSTM solutions were tested and compared in this experiment. See the result session for details.

\vspace{-0.15in}
\paragraph{Data}
In the training process, the face data from the previous experiment was used. One problem is that each face sequence has only 12 time steps which is inconsistent with the 49 time steps in audio sequences. To circumvent this inconsistency, we simply duplicated faces evenly within the 49 time steps. The combinations of face sequences and audio sequences for each identity were randomly paired by the training program during the runtime to maximize the diversity of the training set. For test set of this task, the combinations were however pre-generated. We randomly generated 250,000 correctly paired combinations and 250,000 distractors (ill-paired combinations).

\vspace{-0.15in}
\paragraph{Procedure and Results}
During the comparison, one baseline method and three alternative multimodal LSTM methods were used. In the baseline method, we separately trained a single-modal LSTM only for audio using the same audio data in this experiment. We carefully tuned many hyper parameters, making sure it performed as well as we can achieve. We used it to classify the audio sequence. For face sequence, we used the CNN+SVM approach from the last experiment. Therefore, we shall have 49 proposals for audio labels and another 49 labels for face label proposals. We then looked at the number of labels agreed within these two groups of labels. A threshold $m$ is set to distinguish the sample between distractors and normal samples. For instance, if $m=10$ then the whole multimodal sequence will be classified to distractors if there are more than 10 label proposals temporally disagreed with each other. Otherwise, the multimodal sequence will be classified by averaging the proposals. Note that this distractor rejection procedure was used in all the compared methods in this experiment. The threshold $m$ is tuned to generate various dots in the ROC curve.

The first multimodal solution does not share any weights across the two modality resulting in two separate single-modal LSTMs. We called it ``no cross-modal weight sharing''. The second solution used the weight sharing scheme introduced previously, but did not share $\W_y$ across the modality. Formally, equation (\ref{eq:new_output}) should be re-written as
\begin{eqnarray}
\hspace{-0.1in}y_t^s\hspace{-0.1in} &=& \hspace{-0.1in} softmax(\W_y^s*h_t^s), ~~ s = 1~to~n. \label{eq:new_output_2}
\end{eqnarray}

We called this solution ``half cross-modal weight sharing''. The third solution completely followed the equation (\ref{eq:new_input}) to (\ref{eq:new_output}), named ``full cross-modal weight sharing''.

\begin{figure}[t]
  \centering
  \includegraphics[width=0.9\linewidth]{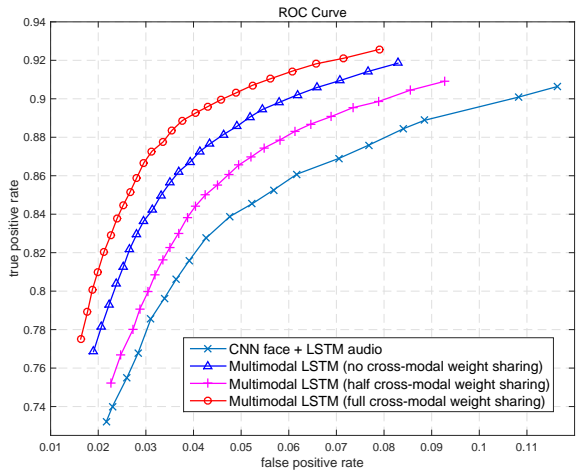}\vspace{-0.1in}
  \caption{Multimodal Long Short-Term Memory.}\vspace{-0.15in}
  \label{fig:fig3} 
\end{figure}

As shown in figure \ref{fig:fig3}, the performance difference is clear. It was expected that the baseline method performed less competitive. However, having isolated $\W_y$ for each modality performed worse than the naive combination of two single-model LSTMs. On the other hand, with the full weight sharing, multimodal LSTM significantly outperforms all other methods.

\vspace{-0.15in}
\paragraph{Discussion}
The false alarm rate was largely increased by not sharing $\W_y$ across modalities. The role of $\W_y$ is to transform the memory cell outputs at each time step to the desirable labels. By sharing this transform across the modality, we can generate more consistent labels for normally paired samples and increased the robustness to distractors. Our experiments showed that this behavior can be automatically captured by the shared $\W_y$.

\subsection{Speaker Identification in The Big Bang Theory}
The last experiment is to apply our method in real-life videos and compare the performance with previous studies.

\vspace{-0.15in}
\paragraph{Data}
To compare with other speaker identification methods, we evaluated the winning multimodal LSTM from the previous experiment in The Big Bang Theory S01E03, as in \cite{Bauml2013,Tapaswi2012,hu2015deep}.

\vspace{-0.15in}
\paragraph{Procedure and Results}
We applied our model to video with the time window of 0.5 second and stride of 0.25 second (e.g. 0s-0.5s, 0.25s-0.75s...). Unlike the controlled setting in the second experiment, the number of distractors in videos varies for each scene. In some cases, there are only distractors in a scene. The evaluation criteria should be more sophisticated. We followed \cite{hu2015deep} to calculate the accuracy of speaker identification to ensure a fair comparison. Specifically, speaker identification is considered successful if a) the speaker is in the scene and the system correctly recognized him/her and correctly rejected all the distractors, or b) the speaker is not in the scene and the system correctly rejected all the distractors in the scene.

Time window more than 0.5 second was also tested to enable a more systematic comparison. We achieved this by further voting within this larger time window. For instance, by having 50\% overlapping of 0.5 second windows in the larger window of 2.0 seconds, we will have seven 0.5 second-sized small windows to vote for the final labels.

Our speaker identification results are reported in table \ref{table:sn-compare}. We compared our method against the state-of-the-art systems. Note that, in \cite{Bauml2013,Tapaswi2012}, as both of them examined the face tracks within the time window specified by the subtitle/transcript segments, they can be viewed as voting on the range of subtitle/transcript segments. As the average time of subtitle/transcript segments in the evaluation video is 2.5s, they are equivalent to our method when evaluated in the voting window of such size. We applied the same voting strategy as in \cite{hu2015deep} under different time window setup. As can be seen from the results, our method outperformed the previous works by a significant margin.

\begin{table}
\begin{center}
\caption{Speaker naming accuracy of different algorithms (\%) in terms of different voting time window (s).}\vspace{-0.1in}
\label{table:sn-compare}
\scalebox{0.82}{
\begin{tabular}{|l||c|c|c|c|c|c|}
\hline
{Time window (s)} & 0.5 & 1.0 & 1.5 & 2.0 & 2.5 & 3.0 \\
\hline\hline
Bauml et al. \citeyear{Bauml2013} & - & - & - & - & 77.81 & - \\
\hline
Tapaswi et al. \citeyear{Tapaswi2012} & - & - & - & - & 80.80 & - \\
\hline
Hu et al. \citeyear{hu2015deep} & 74.93 & 77.24 & 79.35 & 82.12 & 82.81 & 83.42 \\
\hline
Ours & 86.59 & 89.00 & 90.45 & 90.84 & 91.17 & 91.38 \\
\hline
\end{tabular}
}
\end{center}\vspace{-0.15in}
\end{table}

\section{Conclusion}
In this paper, we have introduced a new multimodal LSTM and have applied it to the speaker identification task. The key idea is to utilize the cross-modality weight sharing to capture correlation of two or more temporally coherent modalities. As demonstrated in our experiments, our proposed multimodal LSTM is robust against image degradation and distractors, and has outperformed state-of-the-art techniques in speaker identification. To our knowledge, this is the first attempt in modeling long-term dependencies over multimodal high-level features. We believe our multimodal LSTM is also useful to other applications not limited to the speaker identification task. 

\bibliographystyle{aaai}
\bibliography{mmlstm}

\end{document}